\pdfoutput=1

\documentclass[11pt]{article}

\usepackage[final]{acl}

\usepackage{times}
\usepackage{latexsym}

\usepackage[T1]{fontenc}

\usepackage[utf8]{inputenc}

\usepackage{microtype}

\usepackage{inconsolata}

\usepackage{graphicx}
\usepackage{fullpage}
\usepackage{times}
\usepackage{fancyhdr,graphicx,amsmath,amssymb}
\usepackage{algorithm}
\usepackage{algpseudocode}
\usepackage{booktabs}
\usepackage{adjustbox}
\usepackage{url}
\usepackage{hyperref}
\usepackage{amssymb}
\usepackage{marvosym}
\usepackage{multirow}
\usepackage{subcaption}
\DeclareMathOperator*{\argmax}{arg\,max}

\title{Plug, Play, and Fuse: Zero-Shot Joint Decoding via Word-Level Re-ranking Across Diverse Vocabularies}

\author{Sai Koneru$^{1}$,
   Matthias Huck$^{2}$,
   Miriam Exel$^{2}$, \textnormal{and}
   Jan Niehues$^{1}$ \\
  $^{1}$ Karlsruhe Institute of Technology \\
  $^{2}$ SAP SE, Dietmar-Hopp-Allee 16, 69190 Walldorf, Germany \\
  \texttt{\{sai.koneru, jan.niehues\}@kit.edu} \\
  \texttt{\{matthias.huck, miriam.exel\}@sap.com}}

\begin{document}
\maketitle
\begin{abstract}
\pagenumbering{gobble}
Recent advancements in NLP have resulted in models with specialized strengths, such as processing multimodal inputs or excelling in specific domains. However, real-world tasks, like multimodal translation, often require a combination of these strengths, such as handling both translation and image processing. While individual translation and vision models are powerful, they typically lack the ability to perform both tasks in a single system. Combining these models poses challenges, particularly due to differences in their vocabularies, which limit the effectiveness of traditional ensemble methods to post-generation techniques like N-best list re-ranking. In this work, we propose a novel zero-shot ensembling strategy that allows for the integration of different models during the decoding phase without the need for additional training. Our approach re-ranks beams during decoding by combining scores at the word level, using heuristics to predict when a word is completed. We demonstrate the effectiveness of this method in machine translation scenarios, showing that it enables the generation of translations that are both speech- and image-aware while also improving overall translation quality\footnote{Code can be found at: \url{https://ai4lt.anthropomatik.kit.edu/english/projects_kontextmt.php}}.
\end{abstract}

\section{Introduction}

A broad spectrum of Large Language Models (LLMs) are being developed at an increasing pace, with efforts focused alone or together on adapting them to specific domains \citep{roziere2023code, bolton2024biomedlm, colombo2024saullm}, enhancing their ability to process multiple modalities \citep{liu2023improved, tangsalmonn, li2024llava, beyer2024paligemma}, or training general-purpose LLMs using high-quality data, advanced architectures, and larger numbers of parameters \citep{touvron2023llama, dubey2024llama, jiang2023mistral7b, gemmateam2024gemmaopenmodelsbased}. As a result, numerous models are now publicly available, each with its own unique strengths and weaknesses.

Many use cases, such as image-aware translation in movie subtitling, require combining these strengths because visual cues can be essential for disambiguating the text and ensuring accurate translations.. Currently, LLMs, such as Tower \citep{alves2024tower}, Alma-R \citep{xucontrastive}, and Madlad-400 \citep{kudugunta2024madlad}, excel at translation tasks \citep{kocmi2024preliminary}, while models like PaliGemma \citep{beyer2024paligemma} and LLava \citep{li2024llava} are leading in vision-related tasks. To effectively address image-aware translation, it is essential to harness the strengths of both translation and vision models.

One way to address such a task is to train a multimodal LLM to enhance its translation capabilities without compromising its vision abilities or vice versa. However, this approach requires additional training and task-specific data. Another approach is to leverage ensembling the two models via shallow fusion \citep{gulcehre2015using} or re-ranking the N-best list \citep{hasan2007very}. The disadvantage of shallow fusion is that it assumes both models share the same vocabulary, which is often not the case with current open-source models.

\begin{figure*}[!ht]
\includegraphics[width=\textwidth]{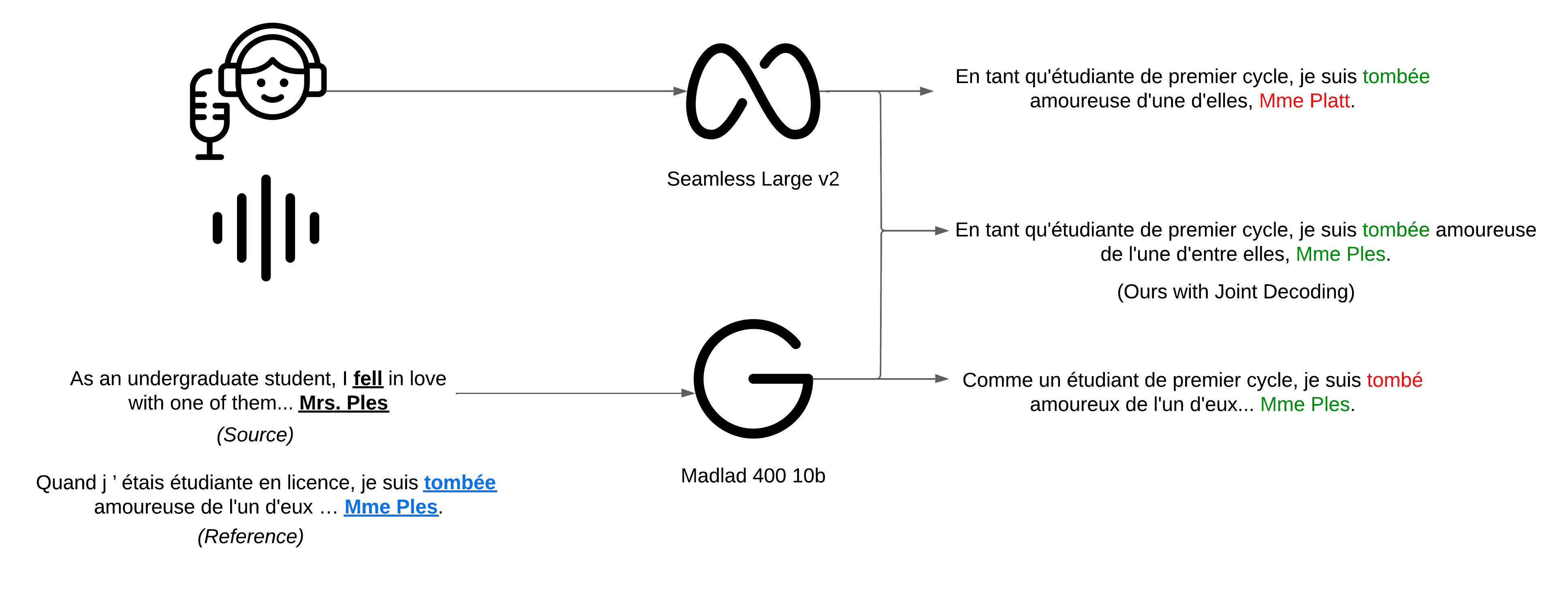}
 \caption{The source sentence to be translated is ambiguous because the translation of the word \textbf{"fell"} can be either masculine \textbf{("tombé")} or feminine \textbf{("tombée")}, depending on the speaker's gender. Seamless-Large V2 \citep{barrault2023seamless} utilizes audio cues to correctly determine the gender form but struggles to accurately translate the name \textbf{"Mrs Ples"} using audio alone. In contrast, the text translation model Madlad-400-10b-mt \citep{kudugunta2024madlad} relies on the gold transcript to correctly translate the name but fails to resolve the gender ambiguity. By combining both models using our approach, the translation correctly captures both the gender form and the named entity.}
\label{fig:mustshe_example}
\end{figure*}

Additionally, re-ranking the N-best list is insufficient because it doesn't allow models to influence each other during decoding. For example, in Figure \ref{fig:mustshe_example}, translating from English to gender-marked language French using audio and transcript shows this limitation. The Speech Translation (ST) model correctly uses the speaker's voice to translate "fell" into the right gender form but misidentifies the name "Ples." On the other hand, the Machine Translation (MT) model correctly translates the name but can't use the speaker's voice for gender disambiguation. Thus, re-ranking falls short, as the correct forms may not even be in the N-best list due to low probability with missing cues.

Furthermore, re-ranking during the decoding process is impractical because the hypotheses are partial and may not align with the tokenization of the ranker model, leading to incorrect probability estimates (Section \ref{sec:tokmismatch}). Thus, resolving vocabulary mismatches by mapping the vocabulary of one model to another \citep{minixhofer2024zero, xu2024bridging} is necessary to allow the merging of probabilities during decoding. However, this approach requires significant additional training steps and can lead to deviations from the original model. Therefore, developing a plug-and-play approach that seamlessly combines different models without requiring additional training or task-specific data is highly advantageous.

This work aims to enable the ranker model to influence the decoding process (online) without any constraints compared to conventional offline N-best list re-ranking. We address this by ensuring that the ranker model only influences the scores for completed words and not for the last word if it is unfinished. Additionally, we propose using the ranker model to determine whether the last word is finished rather than relying on look-ahead approaches to maintain efficiency.

Our main contributions are summarized below:

\begin{enumerate}
    \item \textbf{Online Re-Ranking Algorithm}: We introduce a novel re-ranking algorithm that operates at the word level during decoding at sub-word level, allowing for more accurate tokenization and better integration of information from different models
    \item \textbf{Plug-and-Play Approach}: Our method does not require additional training or task-specific data, making it a flexible and practical solution for integrating multiple models with different strengths.
    \item \textbf{Context-aware Translations}: We demonstrate through experiments including targeted multimodal test sets, which require information from both modalities, that our approach effectively combines the strengths of different models and improves translation quality (Illustrated in Figure \ref{fig:mustshe_example}). 
\end{enumerate}

\section{Methodology}
\label{sec:method}
Given that many models are trained on different tasks, architectures, modalities, and data types, combining these models to leverage inputs from multiple modalities and facilitate knowledge sharing is highly beneficial. Moreover, it is ideal if the ensembling approaches satisfy the following constraints: 1) It should not rely on shared vocabularies for flexibility in choosing models and maximizing potential combinations. 2) Effective knowledge sharing should occur during decoding to better navigate the search space exploiting this knowledge at each step. 3) Avoid requiring additional training, parameters, or major dependence on task-specific data for maximum applicability and not cause deviations from the pre-trained model.  

This section presents our algorithm for ensembling models with different vocabularies that satisfy the aforementioned constraints. First, we explain why re-ranking partial hypotheses can lead to incorrect probability estimates if the word is incomplete. Next, we introduce and justify a heuristic-based approach that predicts whether a hypothesis is at the end of a word, allowing for accurate re-ranking of completed words in partial hypotheses. Finally, we formally describe the complete algorithm, detailing how we merge probabilities from different models and how this process can be integrated with decoding strategies.

\subsection{Challenges of Re-Ranking Partial Hypotheses}
\label{sec:tokmismatch}

Current Neural Machine Translation (NMT) and LLM-based models can utilize various tokenization methods, such as byte-pair encoding (BPE) \citep{sennrich2016neural} or SentencePiece \citep{kudo2018sentencepiece}. These methods often result in distinct vocabularies due to variations in the data and tokenizer training processes. Despite these differences, techniques like re-ranking can still enable estimating the probability of sentences generated from another model. This is achieved by detokenizing the hypothesis from generator model and re-tokenizing it using the ranker model's vocabulary. This process enables the ranker model to produce accurate probability estimates based on its own tokenization scheme.

Now, consider the case of re-ranking while the hypotheses are still being decoded. Assume we have model $\mathcal{M}_{G}$ (the generator) and model $\mathcal{M}_{R}$ (the ranker), each using different tokenizers assign all the tokens in the sentence "Decoding is awesome" with a probability of p for a particular input. However, $\mathcal{M}_{G}$ tokenizes the sentence with subword tokens as "Dec od ing \_is \_awe some," while $\mathcal{M}_{R}$ would tokenize it as "Dec od ing \_is \_awes ome."

If we attempt to re-rank during the decoding process, $\mathcal{M}_{R}$ will provide correct probability estimates up until "\_is" is generated. However, when the generator predicts "\_awe," $\mathcal{M}_{R}$ would incorrectly estimate the probability because it expects "\_awes" instead. Even though both models aim to generate the same sentence, this tokenization mismatch leads to incorrect probability estimates during the decoding process, making online re-ranking challenging.

\subsection{End-of-Word Prediction in Decoding for Accurate Re-Ranking}
\label{sec:endofword}

\begin{algorithm*}[!t]
\caption{Computing merged score of candidate with generator and ranker models.}
\begin{algorithmic}[1]
\setlength{\baselineskip}{1.2em}
\Procedure{MergeScore}{}
    \State \textbf{Input:}   Generator tokens $g_1, g_2, g_3, \dots, g_{n}$, Reranker tokens $r_1, r_2, r_3, \dots, r_{m}$, Generator Model $\mathcal{M}_{G}$, Ranker model $\mathcal{M}_{R}$, Generator Input $\mathcal{I}_{G}$, Ranker Input $\mathcal{I}_{R}$, Re-ranking weight $\alpha$,
    \State \textbf{Output:} $merged\_score$
    \State $next\_tok \gets \argmax \log \mathcal{P}(y| r_1, \dots, r_{m}; \mathcal{I}_{R};\mathcal{M}_{R})$
    \If{$next\_tok[0]$ == "\_" or $next\_tok$ == "<eos>"}
    \State $full_{G} \gets \frac{1}{n}\sum \log \mathcal{P}(g_1, g_2,\dots, g_{n}|\mathcal{I}_{G};\mathcal{M}_{G})$ \Comment{Generator Score for all words}
    \State $full_{R} \gets \frac{1}{m}\sum \log \mathcal{P}(r_1, r_2,\dots, r_{m}|\mathcal{I}_{R};\mathcal{M}_{R})$ \Comment{Ranker Score for all words}
    \State $merged\_score \gets (\alpha) \times full_{G} + (1 - \alpha) \times full_{R}$
    \Else{}
    \State $[g_1,\dots, g_{j}], [g_{j+1},\dots, g_{n}] \gets$ split\_candidate$(g_1,\dots,g_n)$ \Comment{Last word from j+1 token}
    \State $[r_1,\dots, r_{k}], [r_{k+1},\dots, r_{m}] \gets$ split\_candidate$(r_1,\dots,r_m)$ \Comment{Last word from k+1 token}
    \State $prev_G \gets \frac{1}{j}\sum \log \mathcal{P}(g_1, g_2,\dots, g_{j}|\mathcal{I}_{G};\mathcal{M}_{G}) $ \Comment{Generator Score for previous words }
    \State $prev_R \gets \frac{1}{k}\sum \log \mathcal{P}(r_1, r_2,\dots, r_{k}|\mathcal{I}_{R};\mathcal{M}_{R}) $ \Comment{Ranker Score for previous words }
    \State $prev_{GR} \gets (\alpha) \times prev_{G} + (1 - \alpha) \times prev_{R}$
    \State $last_G \gets \sum \log \mathcal{P}(g_{j+1},\dots, g_{n}|\mathcal{I}_{G};\mathcal{M}_{G})$ 
    
    \State $merged\_score \gets  \frac{1}{n}[{prev_{GR} \times j + last_G}]$\Comment{Re-normalized merged score}
    \EndIf
\EndProcedure
\end{algorithmic}
\label{alg:joint}
\end{algorithm*}

While the partially generated hypothesis cannot be accurately ranked at every time step, consider the cases when each word is finished. At that time, we can re-rank the complete hypothesis as the last word is fully generated and the ranker model can tokenize the completed word as it would have done naturally, thereby providing accurate probability estimates. If we know that the last word is incomplete, we can use this information to wait and only rank the previously completed words. Knowing the end of the word enables more precise re-ranking during decoding, even with models that use different tokenization schemes.

Nonetheless, a significant challenge remains: how do we determine when the last word is completed? If the tokenizer places spaces at the right of characters, we could check the predicted token to see if it includes a space, signaling the end of a word. However, this approach is not universal, as many tokenizers do not follow this pattern, and we aim to develop a tokenizer-agnostic solution.

One alternative is to perform a look-ahead step to check if the word has been completed, but this method is also sub-optimal, as it would require decoding twice for each step in the generation process, significantly increasing computational complexity and reducing efficiency. We need a more efficient and generalizable method to determine when a word has been completed during decoding.

To address these challenges, we propose using the ranker model to predict the next token and determine if the word has been completed. This approach offers two key advantages.

Firstly, if the ranker model predicts a space as the next top character, it indicates that the current last word has been completed. The hypothesis will be tokenized correctly, given that it is the prediction from the ranker model itself. Secondly, this prediction can be done together with the re-ranking process by simply also predicting the next token given the previous tokens of the current hypothesis to the ranker model.

This method is more efficient than the look-ahead approach, requiring only one pass of the generator and the ranker model. In contrast, the look-ahead method would require two passes of the generator and one pass of the ranker model. Using the ranker model in this way, we can ensure proper tokenization and accurate probability estimates during the decoding process (online) without additional computational overhead.

\subsection{Integrating Online Re-Ranking with Search}
 
This section formalizes achieving online re-ranking at a word level using beam search as an example of a decoding strategy. Note that the approach can also be applied to other strategies, with slight modifications when necessary. 

A set of candidate sequences is typically maintained during the search, with the number of candidates equal to the configured beam size $b$. At each time step, for each of the $b$ candidate sequences, the model computes likelihood scores for all possible token extensions based on the vocabulary size $V$. This results in a total of $b\times V$ possible extensions. From these $b\times V$ extensions, the top $b$ sequences with the highest scores are selected to form the new set of candidate sequences. This process is repeated iteratively, updating the candidate sequences at each step until enough beams are generated that include end-of-sentence tokens or until a predefined length limit is reached.

To enable re-ranking during the decoding process, we need to adjust the scores of the possible extensions using the ranker model. Directly calculating the likelihood of all extensions would be computationally impractical. Therefore, we introduce a new parameter $topk$, which selects the top $topk$ extensions for each beam during re-ranking.

Hence, at each time step, the generator model calculates the likelihood scores for all $V$ possible extensions for each of the $b$ candidate sequences, resulting in $b\times V$ extensions. Instead of re-ranking all $b\times V$ extensions, the top $topk$ extensions with the highest likelihood scores are selected for each beam. Thus, only $b \times topk$ extensions are considered during re-ranking. For the selected $b\times topk$ extensions, the ranker model estimates their scores and combines them with the original generator scores. For the remaining $b \times (V - topk)$ extensions, the scores are set to $-\infty$ (logically equivalent to discarding them) since they would not be selected in the top beams.

This method significantly reduces computational complexity while allowing effective re-ranking of the most promising candidate extensions, improving the decoding process.

At every decoding step, the problem can be reformulated as determining the merged score of the top candidates according to both models. 

When calculating the merged score during decoding, it's essential to exclude the ranker model's probability if the last word in the current beam is incomplete. This prevents incomplete words from skewing the final score. For beams with incomplete final words, we combine the joint scores of the preceding words with the generator’s score for the last word, ensuring proper normalization to address scale differences between finished and unfinished beams.

After computing the merged scores, we select the top extensions and repeat the process until all beams reach the end-of-sentence token. This method ensures that the final translation is based on fully formed words, optimizing the ranker model's effectiveness and maintaining consistent scoring across all candidates.

\subsubsection{Unified Scoring with Generator and Ranker}

The algorithm to compute the merged score is formally defined in Algorithm \ref{alg:joint} and explained below.

Let us consider two models: the Generator $\mathcal{M}_{G}$ and the Ranker $\mathcal{M}_{R}$. Let $\mathcal{C}$ denote the current candidate for re-ranking and inputs $\mathcal{I}_{G}$ and $\mathcal{I}_{R}$ for $\mathcal{M}_{G}$ and $\mathcal{M}_{R}$ respectively. 

Let the full candidate $\mathcal{C}$ consist of tokens $g_1, g_2, g_3, \dots, g_{n}$ and $r_1, r_2, r_3, \dots, r_{m}$ according to $\mathcal{M}_{G}$ and $\mathcal{M}_{R}$, respectively. Note that $n$ and $m$ denote the length of the sequence, and they may differ due to different tokenization.

The key idea is to rank and merge scores for completed words. We use the ranker model to predict the next token and determine if the last word is finished (\textbf{Line 4}).

\textbf{If the last word is finished}: We can calculate the probability of the full sequence in this case, similar to the case of N-best list re-ranking. First, we calculate the likelihood of the candidate by averaging the log probabilities for both the generator and the ranker (\textbf{Line 6-7}). Then, we merge the scores from both models to determine the final score for the candidate sequence using a hyper-parameter $\alpha$ for weighting (\textbf{Line 8}). This combined score considers the estimates from both models, allowing for contributions from both models.

\textbf{If the last word is incomplete}: We cannot rank the last word due to potential incorrect tokenization. However, we can still estimate the tokens preceding the last word using the ranker model and merge their probabilities. First, we split the candidate into previous and last words based on the ranker and generator (\textbf{Lines 10-11}). We compute the merged score for the previous words using the weighting parameter $\alpha$ (\textbf{Lines 12-14}). For the last word, we rely solely on the generator's scores. To address length normalization issues when combining scores from both models, we re-normalize the merged score for the previous words by multiplying it by the length of the previous word tokens $j$ from the generator, adding the last word's score, and normalizing by the total length $n$ (\textbf{Lines 15-16}).

This integration process ensures that the re-rankers are utilized at the appropriate decoding stages, thereby enhancing the overall quality of the generated sequences by combining the strengths of both models.

\section{Test Suites}

The major advantage of combining models with different vocabularies zero-shot is that it leverages the strengths of available pre-trained models to generate more accurate and robust output. This is particularly relevant in multimodal scenarios, where unimodal systems excel in their respective modalities but are weaker or incapable of processing other modalities. Furthermore, it can also enhance quality compared to N-best list re-ranking when used as an ensembling technique as it waits until the complete sequence is generated. Hence, to validate our approach, we consider three MT scenarios as a test bed where quality can be improved by combining different sources and evaluating with targeted test sets that require information from both models. An overview of test suites is provided in Table \ref{tab:data}.

\subsection{Unimodal MT}

We evaluate the use case of ensembling different LLM models to enhance translation quality. This is particularly relevant given the rapid development of various translation LLMs, where combining different systems can improve quality and robustness. We use the WMT 2022 \textit{English $\rightarrow$ German} test set \citep{kocmi2022findings} to validate our approach and focus solely on assessing translation quality.

\begin{table}[!h]
\resizebox{\columnwidth}{!}{
\begin{tabular}{@{}c|ccc@{}}
\toprule
Test Set & Language Pair                                          & \# Examples                                          & Phenomena                                                                   \\ \midrule
MuST-SHE & \textit{En $\rightarrow$ Fr} & \begin{tabular}[c]{@{}c@{}}315\\ (1108)\end{tabular} & \begin{tabular}[c]{@{}c@{}}Gender Disambiguation\\ Translation\end{tabular} \\ \midrule
CoMMuTE  & \textit{En $\rightarrow$ De} & 300                                                  & \begin{tabular}[c]{@{}c@{}}Word Disambiguation\\ Translation\end{tabular}   \\ \midrule
WMT22    & \textit{En $\rightarrow$ De} & 2037                                                 & Translation                                                                 \\ \bottomrule
\end{tabular}
}
\caption{Overview of test suites. For MuST-SHE, 315 examples are utterances where information is available in audio. However, we use the full test set with other types of bias when reporting translation quality.}
\label{tab:data}
\end{table}

\subsection{Multimodal MT}

Translating from English to gender-marked languages is challenging when the source text lacks clear gender cues. To evaluate bias in current NMT systems, \citet{bentivogli-etal-2020-gender} developed the \textit{MuST-SHE} test suite, which includes examples with varying forms of gender bias. This suite features cases where gender information is conveyed through audio cues, such as the speaker's voice.

While End-to-End ST systems can handle such cases, they often fall short compared to advanced translation LLMs \citep{agarwal2023findings}. Therefore, we use MuST-SHE for \textit{English $\rightarrow$ French} to investigate if combining ST and translation LLMs can improve translation quality and address gender ambiguity.

Similarly, images can assist in disambiguating text and enhancing translation quality. However, translation LLMs typically do not process images, and vision LLMs alone are inadequate for translation tasks. We combine these models to leverage their strengths for better image-aware translations.

Existing vision translation test sets often lack ambiguity, making image inputs unnecessary \citep{vijayan2024case}. To address this, \citet{futeral-etal-2023-tackling} introduced CoMMuTE, which features ambiguous source sentences with two images and their translations. We use CoMMuTE for \textit{English $\rightarrow$ German} translation in a generative framework to evaluate if images can enhance translations without compromising overall quality.

\section{Results}

This section presents the experiments conducted using our ensembling approach across various test suites. Since each test suite has a distinct experimental setup, we will address them individually. First, we will specify the models and evaluation metrics applied in each scenario. Then, we will present the results and highlight our main findings.

\begin{table*}[!t]
\resizebox{2\columnwidth}{!}{
\begin{tabular}{@{}ccc|ccccc@{}}
\toprule
\textit{Generator}     & \textit{Ranker}        & \textit{Online}       & \textit{COMET   22} & \textit{COMET   KIWI 22 QE}                & \textit{COMET   KIWI XXL QE} & \textit{XCOMET-XXL} & \textit{BLEURT} \\ \midrule
                       &                        &                       &                     & \textit{No re-ranking}                     &                              &                     &                 \\ \midrule
GPT-4                  & \textbf{\(\times\)}    & N/A                   & 87.29               & 83.48                                      & 84.91                        & 97.56               & \_              \\
Madlad-10B             & \textbf{\(\times\)}    & N/A                   & 86.60               & 83.14                                      & 82.65                        & 96.77               & 76.79           \\
Alma-13B-R             & \textbf{\(\times\)}    & N/A  & 86.40               & 83.28                                      & 84.25                        & 97.48               & 77.20           \\ \midrule
                       &                        &                       &                     & \textit{Offline re-ranking}                & \textbf{}                    &                     &                 \\ \midrule
Madlad-10B             & Alma-13B-R             & \textbf{\(\times\)}   & 87.27               & 83.68                                      & 84.11                        & 97.12               & 77.66           \\
Madlad-10B, Alma-13B-R & Madlad-10B, Alma-13B-R & \textbf{\(\times\)}   & 87.54               & \textbf{83.95}                             & 84.97                        & 97.39               & 78.20           \\ \midrule
                       &                        &                       &                     & \textit{\textbf{Online re-ranking (ours)}} &                              &                     &                 \\ \midrule
Madlad-10B             & Alma-13B-R             & \textbf{$\checkmark$} & \textbf{87.69}               & 83.94                                      & \textbf{85.20}               & \textbf{97.68}      & \textbf{78.36}  \\ \bottomrule
\end{tabular}
}
\caption{Performance of models on the WMT 22 \textit{English $\rightarrow$ German} test set. Scores are highlighted in \textbf{bold} if it is the best in all configurations. Results for GPT-4 and Alma-13B-R are reported from \citep{xu2024bridging}}.
\label{tab:ensemble}
\end{table*}

\begin{table*}[!ht]
\centering
\resizebox{2\columnwidth}{!}{
\begin{tabular}{@{}cccccc@{}}
\toprule
Model                                                                                 & COMET   22     & COMET   KIWI 22 QE & COMET   KIWI XXL QE & XCOMET-XXL     & BLEURT         \\ \midrule
GPT-4                                                                                 & 87.29          & 83.48              & 84.91               & 97.56          & \_             \\ \midrule
Madlad                                                                                & 86.60          & 83.14              & 82.65               & 96.77          & 76.79          \\ \midrule
(Madlad)  5-best + QE                                                                 & 87.33          & 83.83              & 86.45               & 97.25          & 77.78          \\ \midrule
\begin{tabular}[c]{@{}c@{}}(Madlad + Alma Online \\ re-rank) 5-best + QE\end{tabular} & \textbf{87.66} & \textbf{84.12}     & \textbf{87.86}      & \textbf{97.91} & \textbf{78.31} \\ \bottomrule
\end{tabular}
}
\caption{ Performance of models on the WMT 22 \textit{English $\rightarrow$ German} test set with Quality Estimation based re-ranking via selecting from 5-best list using comet-kiwi-xxl. Scores are highlighted in \textbf{bold} if it is the best in all configurations.}.
\label{tab:qebestlist}
\end{table*}

\subsection{Ensembling for Improving Translations}

\textbf{Models:} We aim to combine two models that excel in translation but possess different strengths. For this purpose, we chose the \textit{Madlad-10B}\footnote{\url{https://huggingface.co/google/madlad400-10b-mt}}, an encoder-decoder architecture trained on extensive parallel data, and \textit{ALMA-13B-R}\footnote{\url{https://huggingface.co/haoranxu/ALMA-13B-R}}, a decoder model trained using contrastive preference optimization and selecting high quality data \citep{xu2024bridging}. 

\textbf{Metrics:} As the models that we would like to ensemble are high quality, we report with several neural metrics to reliably validate the improvements. For reference-based we report with COMET \citep{rei2022comet} and BLUERT \citep{sellam2020bleurt,pu2021learning} whereas for reference-free we report with COMET-KIWI \citep{rei2022cometkiwi}, COMET-KIWI-XXL \citep{rei2023scaling} and XCOMET-XXL \citep{guerreiro2023xcomet} metrics.

\textbf{Hyper-parameters:} We set the re-ranking weight $\alpha$ to $0.5$ given that both models have high quality and should be weighted equally. Furthermore, we set the $topk$ to 5 and the number of beams for the generator as $5$.

To validate our combined model and online re-ranking approach, we compare it against several baselines. First, we check if the ensemble outperforms each individual model. Next, we evaluate if our method surpasses offline re-ranking techniques, indicating a more effective ranker influence and improved search space exploration during decoding.

We evaluate our approach using N-best list re-ranking, with \textit{Madlad} as the generator and \textit{Alma} as the ranker. We generate an N-best list of 25 hypotheses with $\alpha$ set to 0.5 to facilitate a fair comparison between offline and online re-ranking methods. Additionally, we test a scenario where the N-best lists from both models are concatenated and jointly re-ranked on 50 hypotheses.  We reports the results for the baselines and our approach in Table \ref{tab:ensemble}.

\textbf{Ensembling enables to reach state-of-the-art quality:} Both \textit{Madlad} and \textit{Alma} produce high-quality translations, though they still lag behind GPT-4 across all metrics. However, after applying offline re-ranking, their performance improves consistently, becoming competitive with GPT-4. When using our online re-ranking approach, the ensemble outperforms GPT-4 across all metrics and shows our proposed approach can improve the translation quality by a substantial margin.

\textbf{Online re-ranking outperforms offline joint re-ranking:} When \textit{Madlad} serves as the generator and \textit{Alma} as the ranker in our approach, the results are superior to those achieved with joint re-ranking, where both models are used simultaneously. Our approach enhances knowledge sharing and collaboration during the decoding process, leading to better translation quality.

\subsubsection{Quality of N-best list}
The primary motivation behind our approach was to influence the decoding process in real-time, rather than waiting until the end. If this is effective, we expect the N-best list to improve with online re-ranking. Additionally, using quality estimation should enhance the selection of the best hypothesis from the N-best list. To validate this, we utilize COMET-KIWI-XXL for selecting the best candidate from the top 5 beams of \textit{Madlad}, comparing scenarios with and without online re-ranking and report the scores in Table \ref{tab:qebestlist}.

We observe that integrating quality estimation significantly enhances Madlad’s performance across all metrics. Using COMET-KIWI-XXL to select the best candidate from the top 5 beams improves score from $82.65$ $\rightarrow$ $86.45$. This improvement is also evident in the BLUERT score, increasing from $76.79$ $\rightarrow$ $77.78$. Additionally, comparing the top 5 beams with our approach, we find that the quality is superior, demonstrating that the early influence of \textit{ALMA} in decoding. Furthermore, this allows to integration of multiple NMT models to generate the N-best list together and later combined with quality estimation for maximum performance.

\subsection{Speech-Aware Translations}

\textbf{Models:} 
To tackle gender ambiguity in text translation using speaker voice information, we combine a robust text translation model with a speech-based model that excels at disambiguating gender, even if it is not as strong in translation. We use the \textit{Madlad} model \citep{kudugunta2024madlad} for high-quality text translation with gold transcript and the \textit{Seamless}\footnote{\url{https://huggingface.co/facebook/seamless-m4t-v2-large}} model for speech translation. Our approach employs \textit{Madlad} as the \textit{generator} and \textit{Seamless} as the \textit{ranker}, allowing us to leverage the speech model's ability to correct gendered forms in the translation.

However, we observed that the \textit{Seamless} model exhibited a bias toward the masculine gender and struggled to effectively resolve gender ambiguities using speech. To mitigate this, we conducted additional fine-tuning using LoRA \citep{hulora} on a balanced speaker dataset derived from MuST-C (TED talks) with gender annotations \citep{di2019must, gaido2020breeding} (Training details in Appendix \ref{sec:seamless}). We remove talks that are present in MuST-SHE for no overlap. This "debiasing" process improved the model's ability to disambiguate gender based on speech. Consequently, we use the \textit{Madlad} and adapted \textit{Seamless} models to generate high-quality, speech-aware translations.

\begin{table*}[!ht]
\resizebox{2\columnwidth}{!}{
\begin{tabular}{@{}cccccccc|cc@{}}
\toprule
\multicolumn{1}{c|}{\textit{Generator}}   & \textit{Ranker}      & \multicolumn{1}{c|}{\textit{Online}}       & \textit{\begin{tabular}[c]{@{}c@{}}1F\\ (Acc \%)\end{tabular}} & \textit{\begin{tabular}[c]{@{}c@{}}1F\\ (Term Cov \%)\end{tabular}} & \textit{\begin{tabular}[c]{@{}c@{}}1M\\ (Acc \%)\end{tabular}} & \textit{\begin{tabular}[c]{@{}c@{}}1M\\ (Term Cov \%)\end{tabular}} & \textit{\begin{tabular}[c]{@{}c@{}}Avg\\ (Acc \%)\end{tabular}} & \multicolumn{2}{c}{\textit{COMET}}      \\
\multicolumn{1}{c|}{}                     &                      & \multicolumn{1}{c|}{}                      &                                                                &                                                                     &                                                                &                                                                     &                                                                 & \textit{Correct} & \textit{$\triangle$} \\ \midrule
                                          &                      &                                            &                                                                & \textit{No re-ranking}                                              &                                                                &                                                                     &                                                                 &                  &                      \\ \midrule
\multicolumn{1}{c|}{Madlad}               & \textbf{\(\times\)}  & \multicolumn{1}{c|}{N/A}                   & 25.92                                                          & 68.39*                                                              & 90.44*                                                         & 63.65*                                                               & 58.18                                                           & 83.52             & 0.90                  \\
\multicolumn{1}{c|}{Seamless}             & \textbf{\(\times\)}  & \multicolumn{1}{c|}{N/A}                   & 20.28                                                          & 63.20                                                               & 88.30                                                          & 62.43                                                               & 54.29                                                           & 79.31             & 0.73                  \\
\multicolumn{1}{c|}{Seamless Bal}         & \textbf{\(\times\)}  & \multicolumn{1}{c|}{N/A}                   & 50.18 *                                                        & 62.73                                                               & 65.89                                                          & 59.02                                                      & 58.03                                                           & 80.48            & 0.83                 \\ \midrule
                                          &                      &                                            &                                                                & \textit{Offline re-ranking}                                         & \textbf{}                                                      &                                                                     &                                                                 &                  &                      \\ \midrule
\multicolumn{1}{c|}{Madlad}               & Seamless Bal         & \multicolumn{1}{c|}{\textbf{\(\times\)}}   & 28.81                                                          & 67.92                                                               & \textbf{89.59}                                                 & 63.41                                                      & 59.20                                                           & 83.66            & 0.96                 \\
\multicolumn{1}{c|}{Seamless Bal}         & Madlad               & \multicolumn{1}{c|}{\textbf{\(\times\)}}   & \textbf{40.90}                                                 & 65.09                                                               & 77.99                                                          & 60.97                                                      & 59.44                                                           & 81.31            & 0.90                 \\
\multicolumn{1}{c|}{Madlad, Seamless Bal} & Madlad, Seamless Bal & \multicolumn{1}{c|}{\textbf{\(\times\)}}   & 29.83                                                          & 67.92                                                               & \textbf{89.59}                                                 & 63.41                                                               & 59.71                                                           & 83.64            & 0.96                 \\ \midrule
                                          &                      &                                            &                                                                & \textit{\textbf{Online re-ranking (ours)}}                          &                                                                &                                                                     &                                                                 &                  &                      \\ \midrule
\multicolumn{1}{c|}{Madlad}               & Seamless Bal         & \multicolumn{1}{c|}{\textbf{$\checkmark$}} & 33.78                                                          & \textbf{68.16}                                                      & 86.86                                                          & \textbf{63.65*}                                                     & \textbf{60.32*}                                                 & \textbf{83.78*}  & \textbf{1.1*}        \\ \bottomrule
\end{tabular}
}
\caption{Performance of models on the MuST-SHE test set for speech-aware translations. \textit{Seamless Bal} indicates the adapted model trained on balanced gender data. $\triangle$ denotes the \textit{sensitivity}, i.e., the difference in scores between correct and incorrect references. Scores are highlighted in \textbf{bold} if online re-ranking improves over offline re-ranking and \textasteriskcentered{} if it is the best in all configurations.}
\label{tab:speechaware}
\end{table*}

\textbf{Metrics:} To evaluate the effectiveness of our approach in disambiguating gender and improving translation quality, we use several key metrics. For gender disambiguation, we follow the methodology of \citet{bentivogli-etal-2020-gender} and report two metrics: accuracy (correct gender form is present) and coverage (either gender form is present). 


For overall translation quality, we report BLEU \citep{papineni2002bleu}, ChrF2 \citep{popovic2016chrf} calculated using SacreBLEU \citep{post2018call}, and COMET \citep{rei2022comet} (\textit{wmt22-comet-da}) for brevity.

Additionally, we report \textit{Sensitivity}, which measures the difference between the scores of correctly and incorrectly gendered references, as suggested by \citet{bentivogli-etal-2020-gender}. 


\begin{table*}[!ht]
\centering
\resizebox{1.7\columnwidth}{!}{
\begin{tabular}{@{}ccc|cccccc@{}}
\toprule
\multirow{2}{*}{\textit{Generator}} & \multirow{2}{*}{\textit{Ranker}} & \multirow{2}{*}{\textit{Online}} & \multicolumn{2}{c}{\textit{BLEU}}       & \multicolumn{2}{c}{\textit{Chrf2}}      & \multicolumn{2}{c}{\textit{COMET}}      \\ \cmidrule(l){4-9} 
                                    &                                  &                                  & \textit{Correct} & \textit{$\triangle$} & \textit{Correct} & \textit{$\triangle$} & \textit{Correct} & \textit{$\triangle$} \\ \midrule
Madlad-10B                          & \textbf{\(\times\)}              & N/A                              & 45.9             & 0.4                  & 62.3             & 1.3                  & 82.90            & 0.06                 \\
PaliGemma-3B MT                     & \textbf{\(\times\)}              & N/A              & 27.6             & \textbf{5.7}         & 51.0             & \textbf{7.3}         & 79.58            & \textbf{8.25}        \\
Madlad-10B                          & PaliGemma-3B MT                  & \textbf{\(\times\)}              & 46.1             & 1.9                  & \textbf{62.6}    & 1.7                  & \textbf{83.45}   & 1.17                 \\
Madlad-10B                          & PaliGemma-3B MT                  & \textbf{$\checkmark$}            & \textbf{46.2}    & 1.8                  & \textbf{62.6}    & 1.9                  & 83.25            & \textbf{1.34}                 \\ \bottomrule
\end{tabular}
}
\caption{ Performance of models on the CoMMuTE \textit{English $\rightarrow$ German} test set for image-aware translations. $\triangle$ indicates the sensitivity i.e difference between correct and incorrect references. Scores are highlighted in \textbf{bold} if it is the best in all configurations.}.
\label{tab:imageaware}
\end{table*}

\textbf{Hyper-parameters:} 
For decoding with \textit{Madlad}, we use beam search with 5 beams. Our proposed algorithm involves two key parameters: $\alpha$ and $topk$. We set $topk$ to 5, resulting in a total of 25 candidates being ranked by \textit{Seamless} at each step.

We optimized $\alpha$ through grid search on the MuST-C development set (Appendix \ref{sec:hyper}) via offline re-ranking and setting it to 0.8 based on these results. We also create an N-best list of 25 hypotheses with $\alpha$ at 0.8 for offline comparison and perform joint re-ranking on the combined 50 N-best lists. Results are summarized in in Table \ref{tab:speechaware}.

\textbf{\textit{Madlad} and \textit{Seamless} complement each other:} \textit{Madlad} excels in overall translation quality (83.5) compared to \textit{Seamless} (79.31). While \textit{Seamless} initially favors masculine terms, fine-tuning on balanced data improves overall quality to 80.48, significantly reducing masculine bias (90.44 to 65.89) and increasing feminine representation (25.92 to 50.18). Thus, the adapted \textit{Seamless} demonstrates improved gender disambiguation, though \textit{Madlad} remains superior in overall translation. Hence, combining the models can be highly beneficial.

\textbf{Online re-ranking improves overall translation quality:} After re-ranking with N-best list, we see that the translation quality is improved when \textit{Madlad} as a generator and \textit{Seamless Bal} as a ranker model ($83.50 \rightarrow 83.66$). In the opposite scenario where \textit{Seamless Bal} uses \textit{Madlad} as a ranker model, the quality also improves ($80.48 \rightarrow 81.31$) but is lower than \textit{Madlad} alone. However, during online re-ranking, we see that we achieved the best performance of $83.78$. This suggests that our approach facilitates knowledge sharing between the models during decoding, leading to significant quality enhancements.

\textbf{Balance between translation quality and gender disambiguation through online re-ranking} We observe that the highest accuracies for feminine terms (1F) are achieved when \textit{Seamless Bal} is employed as a generator. Nevertheless, the overall translation quality in these instances is considerably lower compared to scenarios where \textit{Madlad} is the generator. By using \textit{Madlad} as a generator, we attain a higher average 1F score of 60.32 compared to offline re-ranking without compromising overall translation quality and better distribution across gender. Moreover, we achieved the highest sensitivity score of $1.1$ across all configurations. This shows that our approach can consistently perform better than traditional N-best list re-ranking.

While the scores for the disambiguation are not high, we would like to highlight that we focused on combining the strengths of the models. However, one can use targeted systems such as \citet{gaido2020breeding} to further improve the performance for the desired tasks.

\subsection{Image-Aware Translations}

\textbf{Models:} To integrate image information for disambiguating source text, a robust multimodal machine translation (MT) system is essential. Initially, we experimented with the off-the-shelf instruction-tuned Llava model\footnote{\url{https://huggingface.co/llava-hf/llava-v1.6-vicuna-13b-hf}} \citep{li2024llava}. While Llava provided reasonable results, its performance was sub-par for our needs. Consequently, we chose to fine-tune the \textit{PaliGemma} model\footnote{\url{https://huggingface.co/google/paligemma-3b-ft-cococap-448}} \citep{beyer2024paligemma}, which was originally trained to generate captions in multiple languages. We fine-tuned \textit{PaliGemma} using the Multi30k image captions dataset \citep{W16-3210}, adapting it with Q-LoRA (Appendix \ref{sec:paligemma}) for enhanced image-aware translations (\textit{PaliGemma-3B MT}).

\textbf{Metrics:} For evaluating this task, we use BLEU, ChrF2, and COMET scores, as we do not have specific annotations for words in the target sentences. To assess the impact of contextual information provided by the images, we also report the sensitivity metric $\triangle$, to estimate how much the image context influences the translation quality.

\textbf{Hyper-parameters:} Vision LLMs require more memory because the image is encoded into a long sequence of tokens. Consequently, we were limited to using a beam size of 3 with a top-k of 3. Additionally, tuning the parameter $\alpha$ was challenging due to the lack of a dedicated ambiguous test set; using a standard test set would result in no weight being given to the vision model. Therefore, we report the oracle $\alpha$ of $0.9$, which represents the best-performing weight on the test set, determined through a grid search with offline re-ranking. We report the scores in Table \ref{tab:imageaware}. 

\textbf{PaliGemma is highly sensitive to image context:} We observe that the sensitivity $\triangle$ of our fine-tuned \textit{PaliGemma} model for MT is notably high across all metrics (e.g., 5.7 BLEU), demonstrating that the model is effectively using the image information to influence its translations. This suggests that \textit{PaliGemma} does not disregard the visual context during translation. However, despite this sensitivity, \textit{PaliGemma's} overall translation quality significantly lags behind that of \textit{Madlad}, as indicated by the lower COMET score (difference of 3.32). This disparity highlights the potential benefit of combining the strengths of both models to achieve more accurate and image-aware translations.

\textbf{No clear winner between offline and online re-ranking:} Comparing offline and online re-ranking, we find that re-ranking with \textit{PaliGemma} enhances translations, evidenced by a sensitivity $\triangle$ increase of up to 1.28 COMET. There's also a slight improvement in overall translation quality after re-ranking. However, the difference between the two approaches is modest, especially given the small test set size of 300 examples.

We hypothesize two main factors behind the results. First, \textit{Madlad} assigns very low probabilities to translations of ambiguous words it isn't biased toward, while \textit{PaliGemma} avoids extremely high probabilities. As a result, merging probabilities tends to favor the incorrect translation with the highest overall score. Second, the test sentences are short, averaging 4-5 words, so the N-best list includes diverse variations, making offline re-ranking similar to the online approach. However, we believe our online re-ranking method could benefit longer sentences and stronger vision translation models.
\section{Related Work}
\textbf{Fusion for MT:} Integrating additional language models into MT systems via shallow or deep fusion, or through re-ranking, to improve translation quality is a well-studied area \citep{chen2006itc, hasan2007very, gulcehre2015using, li2016mutual, gulcehre2017integrating, herold2023improving}. \citet{stahlberg2018simple} explored advanced fusion method where an NMT model is trained from scratch while keeping a pre-trained language model fixed, allowing the model to learn only what is missing. There has also been growing interest in combining NMT with document-level language models \citep{stahlberg2019cued, petrick-etal-2023-document, hoang2024fly}. Unlike previous works that utilize static weights for merging probabilities, \citet{jean2020log} propose dynamic coefficients, which are crucial for effectively combining models with different strengths. 

\textbf{Ensembling:} System combination, which involves merging multiple hypotheses to generate a better version, is one approach to leveraging the strengths of different models \citep{bangalore2001computing, matusov2006computing, heafield2010combining, freitag2014jane}. Another approach is to merge model parameters \citep{junczys2016amu} or distill knowledge from the models \citep{freitag2017ensemble}. With the increasing diversity of LLMs, recent research has explored methods to combine them through vocabulary merging \citep{xu2024bridging}, generating new outputs based on hypotheses \citep{jiang2023llm}, or dynamically selecting different models at each step \citep{shen2024learning}.

Our work differs from these approaches as it neither relies on vocabulary matching nor requires additional training data.

\section{Conclusion}

We proposed a novel ensembling strategy that operates at the word level during the decoding process to enhance knowledge sharing. Our approach demonstrated significant benefits across multiple scenarios. It proved effective for ensembling translation systems, and even when combined with quality estimation models, it achieved state-of-the-art translation quality. Additionally, experiments on targeted multimodal test sets revealed that our method facilitates better knowledge sharing compared to traditional re-ranking techniques.

For future work, we propose to explore unsupervised dynamic selection, enabling models to generate outputs only when they are better equipped for the task. We believe this approach could address the current limitations and lead to more significant improvements in image-aware translation.

\section{Limitations}
The major limitation of this work is that we operate at word-level which is not compatible for several languages that are character based. Hence, it is not trivial to merge models for generating such languages. Further analysis is necessary on character-level tokenization to accurately re-rank during the decoding steps.

Another drawback is that, although re-ranking enhances translation quality, it incurs a latency cost. Unlike offline re-ranking, our approach employs the ranker model at each time step, resulting in significantly slower performance.

Finally, we focused mainly on ensembling the two models using static weights. However, since the models have different strengths, it is crucial to determine when to rely on one model or ensemble both. This dynamic approach would better exploit each model's strengths while avoiding the integration of their weaknesses.

\bibliography{custom}

\appendix

\section{Appendix}

\subsection{Adapting Seamless}
\label{sec:seamless}

We use the gender annotations from \citet{gaido2020breeding} to select talks with feminine speaker pronouns and an equal amount of randomly sampled masculine talks that are in the training set. We use the huggingface transformer's library \citep{wolf2019huggingface} for fine-tuning Seamless. We use LoRA \citep{hulora} to fine-tune Seamless on this data. We set the \textit{rank} to $16$, \textit{lora\_alpha} to $64$ and \textit{lora\_dropout} to $0.1$. We apply adapters on the following modules: \textit{q\_proj, v\_proj, linear\_q, linear\_v}. We set \textit{batch\_size} to $16$, \textit{gradient\_accumulation\_steps} to $8$ and train with \textit{fp16} for $20$ epochs validating at every $200$ steps. The \textit{learning\_rate} is set to $1e^{-5}$. The other parameters are set to default in the transformers library.

\subsection{Adapting PaliGemma}
\label{sec:paligemma}

We also fine-tune the PaliGemma model with the huggingface transformer's library \citep{wolf2019huggingface} but use Q-LoRA \citep{dettmers2023qlora} with \textit{4-bit} quantization as the vision models require more VRAM. We set the \textit{rank} to $8$, \textit{lora\_alpha} and \textit{lora\_dropout} to default. We apply adapters on the following modules: \textit{q\_proj, k\_proj,  v\_proj, gate\_proj, up\_proj, down\_proj}. We set \textit{batch\_size} to $2$, \textit{gradient\_accumulation\_steps} to $6$ and train with \textit{bf16} for $5$ epochs validating at every $200$ steps. The \textit{learning\_rate} is set to $2e^{-5}$ with \textit{AdamW} optimizer.  The other parameters are set to default in the transformers library.

\subsection{Hyper-parameter Tuning for Speech-Aware Translations}
\label{sec:hyper}

\begin{figure*}
     \centering
     \begin{subfigure}[b]{0.6\textwidth}
         \centering
         \includegraphics[width=\textwidth]{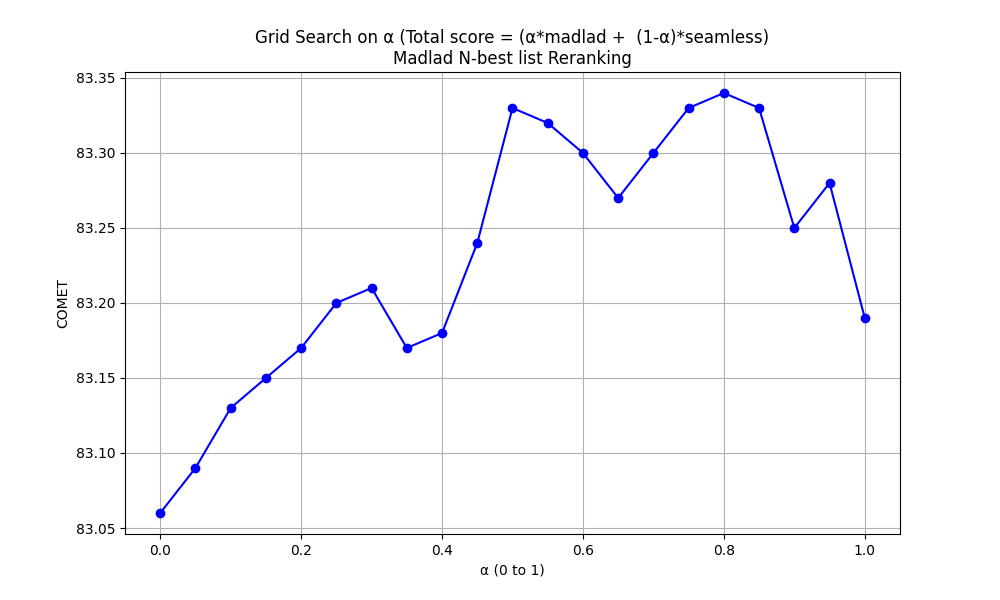}
         \caption{Re-ranking with Madlad N-best list}
         \label{fig:madlad}
     \end{subfigure}
     \hfill
     \begin{subfigure}[b]{0.6\textwidth}
         \centering
         \includegraphics[width=\textwidth]{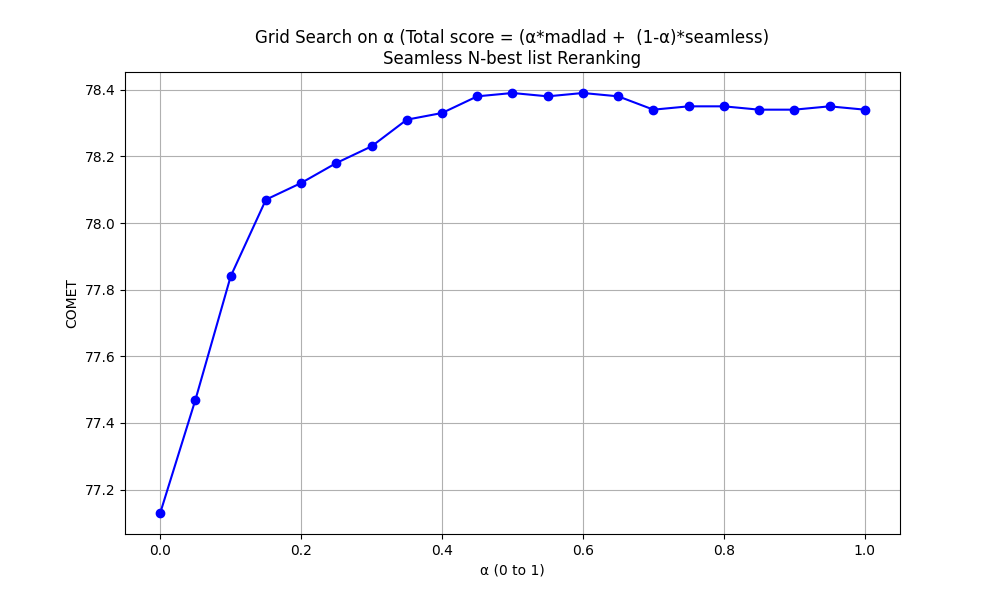}
         \caption{Re-ranking with Seamless N-best list}
         \label{fig:seamless}
     \end{subfigure}
     \hfill
     \begin{subfigure}[b]{0.6\textwidth}
         \centering
         \includegraphics[width=\textwidth]{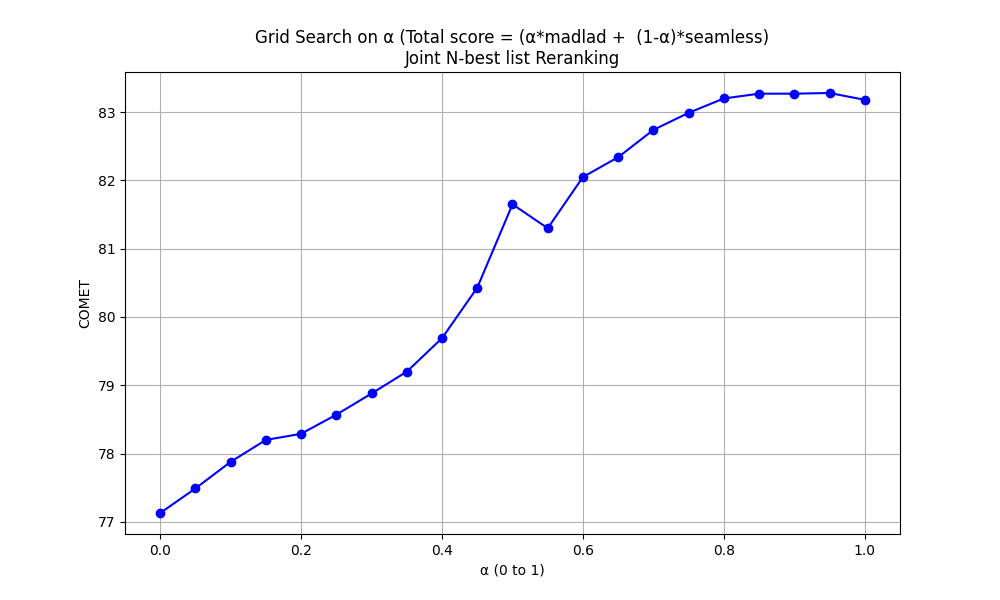}
         \caption{Re-ranking with Joint N-best list}
         \label{fig:joint}
     \end{subfigure}
        \caption{Grid Search on $\alpha$ with \textit{Madlad} and \textit{Seamless Bal} on the MuST-C development set with N-best lists from different generators and rankers.}
        \label{fig:three graphs}
\end{figure*}

To find the re-ranking weight $\alpha$, we generate the 25-best list of \textit{Madlad} and \textit{Seamless} on the MuST-C development set. Then, we calculate the scores of the models on these hypothesis and perform a grid search to find the optimal weight. Here, $\alpha=1$ means that the score is only from \textit{Madlad} and $\alpha=0.5$ means equal contribution. The grid search is plotted in Figure \ref{fig:three graphs}. 

We see that $\alpha$ as $0.8$ is always achieving higher scores. Furthermore, we see that using \textit{Seamless} as generator (Figure \ref{fig:seamless}) leads to poor translation quality and $\alpha$ as $1$. However, in the case of \textit{Madlad} as a generator (Figure \ref{fig:madlad}), we see that $\alpha$ as $1$ is not optimal showing that re-ranking with \textit{Seamless} is indeed beneficial. Finally in the case of both models as generator (Figure \ref{fig:joint}), we again see that $\alpha$ as $1$ achieves highest quality showing that \textit{Seamless} is not beneficial.

\end{document}